\documentclass[letterpaper]{article}
\usepackage{aaai19}
\usepackage{times}
\usepackage{helvet}
\usepackage{courier}
\usepackage[hyphens]{url} 
\usepackage{graphicx}
\usepackage{graphicx}
\usepackage{xparse}
\usepackage{amsmath}
\usepackage{amsthm}
\usepackage{amssymb}
\usepackage{xspace}
\usepackage{relsize}
\usepackage{multirow}
\usepackage{comment}

\newcommand{\subsubparagraph}[1]{}

\newtheorem{defi}{Definition}

\makeatletter
\let\@myref\ref

\newcommand{\refsec}[1]{Sec.\,\@myref{#1}}
\newcommand{\refseq}[1]{Sec.\,\@myref{#1}}
\newcommand{\refig}[1]{Fig.\,\@myref{#1}}
\newcommand{\refigs}[2]{Fig.\,\@myref{#1}-\@myref{#2}}

\newcommand{\reftbl}[1]{Table \@myref{#1}}
\newcommand{\refstep}[1]{Step \@myref{#1}}
\newcommand{\refalgo}[1]{Algorithm \@myref{#1}}
\newcommand{\refchap}[1]{Chapter \@myref{#1}}
\newcommand{\reflst}[1]{List \@myref{#1}}
\newcommand{\refeq}[1]{Eq. \@myref{#1}}

\newcounter{list}[section]

\makeatother

\newcommand{\brackets}[1]{{\left<#1\right>}}
\newcommand{\braces}[1]{{\left\{#1\right\}}}
\newcommand{\parens}[1]{{\left(#1\right)}}

\newcommand{\uline}[1]{\underline{#1}}

\newcommand{\astar}{\xspace {$A^*$}\xspace}

\newcommand{\defun}[1]{%
\makeatletter
\expandafter\def\csname the#1\endcsname{\text{\it #1}}
\expandafter\def\csname #1\endcsname ##1{\csname the#1\endcsname\left(##1\right)}%
\makeatother
}

\newcommand{\defsetop}[2]{%
\makeatletter
 \expandafter\def\csname #1\endcsname ##1##2##3{%
  \expandafter\def\csname #1arg\endcsname{##1}%
  \expandafter\def\csname #1set\endcsname{##2}%
  \expandafter\def\csname #1cond\endcsname{##3}%
  \braces{##1##2\mid #2 ##3}%
 }%
\makeatother%
}

\defun{precond}
\defun{condition}
\defun{params}
\defun{type}
\defun{proc}
\defun{cost}

\defun{push}
\defun{pref}

\defun{init}
\defun{goal}
\defun{objects}
\defun{task}

\defun{parent}
\defun{plateau}

\defsetop{filter}{}
\defsetop{map}{}

\makeatletter
\NewDocumentCommand{\todo}{s O{} m}{%
  \IfBooleanTF#1%
    {\@todos{#2}{#3}}%
    {\@todons{#2}{#3}}}
\newcommand{\@todons}[2]{}
\newcommand{\@todos}[2]{}
\makeatother

\def\_{\\[-0.3em]}

\makeatletter

\newcommand{\newheuristic}[2]{%
 \def#1{%
  \ifmmode%
  h^\text{#2}\xspace%
  \else%
  \text{#2}\xspace%
  \fi%
 }%
}

\newheuristic{\lmcut}{LMcut}
\newheuristic{\mands}{M\&S}
\newheuristic{\pdb}{PDB}
\newheuristic{\ff}{FF}
\newheuristic{\ce}{CEA}
\newheuristic{\cg}{CG}
\newheuristic{\ad}{add}
\newheuristic{\lc}{LC}

\newcommand{\newUnitCostHeuristic}[2]{%
 \def#1{%
  \ifmmode%
  \hat{h}^\text{#2}\xspace%
  \else%
  \text{#2}\xspace%
  \fi%
 }%
}

\newUnitCostHeuristic{\lmcuto}{LMcut}
\newUnitCostHeuristic{\mandso}{M\&S}
\newUnitCostHeuristic{\ffo}{FF}
\newUnitCostHeuristic{\ceo}{CEA}
\newUnitCostHeuristic{\cgo}{CG}
\newUnitCostHeuristic{\ado}{add}
\newUnitCostHeuristic{\gco}{GoalCount}
\newUnitCostHeuristic{\lco}{LC}

\makeatother

\def\latentplanner{Latplan\xspace}

\newcommand{\before}{pre}
\newcommand{\after}{suc}

\let\tb\textbf

\author{
Masataro Asai \\ IBM Research \\ MIT-IBM Watson AI Lab \And
Hiroshi Kajino \\ IBM Research}
\title{Towards Stable Symbol Grounding\\ with Zero-Suppressed State AutoEncoder}

\begin{document}
\maketitle
\begin{abstract}
While classical planning has been an active branch of AI,
its applicability is limited to the tasks precisely modeled by humans.
Fully automated high-level agents should be instead able to find a symbolic representation
of an unknown environment without supervision,
otherwise it exhibits the knowledge acquisition bottleneck.
Meanwhile, \latentplanner \cite{Asai2018} partially resolves the bottleneck
with a neural network
called State AutoEncoder (SAE). SAE obtains the propositional
representation of the image-based puzzle domains with unsupervised learning,
generates a state space and performs classical planning.
In this paper,
we identify the problematic, stochastic behavior of the SAE-produced propositions 
as a new sub-problem of symbol grounding problem, the symbol \emph{stability} problem.
Informally, symbols are
\emph{stable} when their referents (e.g. propositional values) do not change against small perturbation of the observation,
and unstable symbols are harmful for symbolic reasoning.
We analyze the problem in Latplan both formally and empirically, and
propose ``Zero-Suppressed SAE'', an enhancement that stabilizes the propositions
using the idea of closed-world assumption as a prior for NN optimization.
We show that
it finds the more stable propositions and the more compact representations, resulting in an improved success rate of \latentplanner.
It is robust against various hyperparameters and eases the tuning effort, and also provides a weight pruning capability as a side effect.
\end{abstract}

\section{Introduction}

Symbol grounding problem \cite{harnad1990symbol,Steels2008} is one of the key milestones in AI research
which seeks to achieve high-level intelligence.
In Physical Symbol Systems Hypothesis \cite{newell1976computer}, it is believed that
an agent with high-level intelligence performs tasks by efficiently manipulating a compact set of abstract symbols.
Symbolic manipulation
allows for
the development of highly optimized and generalized, domain-independent heuristics \cite{Hoffmann01,Helmert2009}
that can be easily applied to multiple tasks with few or no data,
while the current learning-based approaches struggle to improve its multi-task performance and data efficiency.
To enable such a symbolic computation in a real-world environment,
agents should be able to find the symbolic representation of the environment by itself.

Recently, Latplan system \cite{Asai2018} successfully
connected a subsymbolic neural network (NN) system and a symbolic Classical Planning system
to solve various visually presented puzzle domains.
The State AutoEncoder (SAE) neural network in \latentplanner
generates a set of propositional symbols from the training images with no additional information
and provides a bidirectional mapping between images and propositional states.
The system then solves the propositional planning problem using a classical planner Fast Downward \cite{Helmert04}
and returns an image sequence that solves the puzzle
by decoding the intermediate propositional states of the plan.
It also discovers a set of action symbols that distinguish the modes of
state transitions through AMA$_2$ unsupervised learning process.
Thus the system grounds two kinds of symbols:
Propositional symbols and action symbols,
and opens a promising direction for applying a variety of symbolic methods to the real world.
The search space generated by \latentplanner was shown to be compatible
to a symbolic Goal Recognition system \cite{amado2018goal}.
Another approach replacing SAE/AMA$_2$ with InfoGAN was also proposed recently \cite{kurutach2018learning}.

Despite its success,
the propositional representations learned by SAEs have a problematic behavior
due to its lack of strong guarantees on the learned results.
That is, while the SAE can reconstruct the input with high accuracy,
the learned latent representations are not ``stable'', i.e.,
some propositions may flip the value (true/false) randomly
given the identical or nearly identical image input.
This is mainly because SAE learns the mapping between images and binary representations as a many-to-many relationship.
While this property has not been considered as an issue in the machine learning community where only the accuracy matters,
its unstable latent representation poses a significant threat to the reasoning ability of the planners.

Unstable symbols are harmful for symbolic reasoning because
they break the identity assumption built into the reasoning algorithms.
For instance, in \latentplanner, 
a single image may map to multiple propositional states due to stochasticity;
therefore the duplication detection in search algorithms such as \astar \cite{hart1968formal}
fails to realize that a single real-world state is visited multiple times through
different symbolic state representations.
Unlike machine learning tasks,
the symbolic planning requires a mapping
that abstracts many images into a single symbolic state, i.e., many-to-one mapping.
To this end, it is necessary to develop an autoencoder that learns a
many-to-one relationship between images and binary representations.

\begin{figure}[tb]
 \centering
 \includegraphics{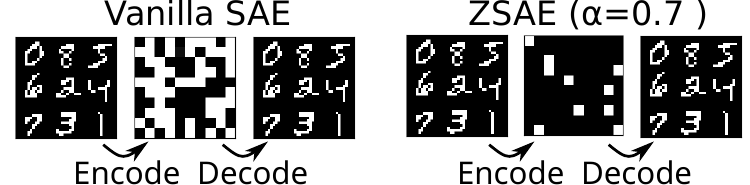}
 \caption{
Autoencoding results of an MNIST 8-puzzle state
using a vanilla State AutoEncoder (SAE) \cite{Asai2018} and a proposed Zero-Suppressed SAE (ZSAE) with 100 propositions.
ZSAE obtains a compact representation that uses fewer true bits.
}
 \label{zsae-overview}
\end{figure}

The contribution of this paper is threefold.
The first contribution of this paper is the identification of a sub-problem of symbol grounding
called ``symbol \emph{stability} problem'' (SSP), which seeks to find a set of symbols
whose values/referents stays the same for the same/similar raw inputs.
Stability is orthogonal to the \emph{performance/accuracy} of NNs:
NNs are known to predict the correct output robustly, but they do not guarantee
the quality of the internal representation.
It is also unrelated to the numerical stability of the training or the reproducibility of the results.

The second contribution of this paper is the formal analysis of the Latplan system,
in particular, the comparison to the original mathematical model of Gumbel-Softmax
\cite{jang2016categorical}.
The analysis revealed that Latplan's Gumbel-Softmax has a deviation from
the original, which was the key to its first success: The slight
modification acted as an \emph{Entropy Regularization} term for the SAE neural
network that suppresses the randomness to some extent.

The third contribution of this paper is
the proposal of Zero-Suppressed State AutoEncoder (ZSAE, \refig{zsae-overview}).
Inspired by the fact that the \emph{Entropy Regularization} stabilizes the representation,
ZSAE further stabilizes the propositions 
by an additional regularization term which
guides the network optimization so that unused propositions tend to 
take the value of zero (false) instead of random values.
The stable representation results in a higher success rate of classical planning.
Also, the network is less sensitive to the network size (hyperparameters)
as it automatically reduces the number of bits used.
Moreover, we show that we can reduce the memory usage of the network
by pruning some unused neurons
that now have a constant activation of zero instead of random values.

\section{Preliminaries}
\label{background}

\textbf{Symbol grounding} is an unsupervised process of establishing a mapping
from huge, noisy, continuous, unstructured inputs
to a set of compact, 
discrete, identifiable (structured) entities, i.e., symbols \cite{harnad1990symbol,Steels2008,Asai2018}.
PDDL \cite{McDermott00} has six kinds of symbols: Objects, predicates, propositions, actions, problems and domains.
Each type of symbol requires its own mechanism for grounding.
For example, the large body of work in the image processing community on recognizing 
objects (e.g., faces) and their attributes (male, female) in images, or scenes in videos (e.g., cooking)
can be viewed as corresponding to grounding the object, predicate and action symbols, respectively.
In this paper, we focus on grounding the propositional symbols.

\textbf{\latentplanner} \cite{Asai2018} is a framework for
\emph{domain-independent image-based classical planning}.
\latentplanner is able to ground
the propositional and action symbols.
Classical planners such as FF \cite{Hoffmann01} or
FastDownward \cite{Helmert04} takes a PDDL model as an input, which
specifies the state representation and the transition rules.
In contrast, \latentplanner learns the state representation as well as the transition rules
entirely from the image-based observation of the environment with deep NNs.
The system was shown to solve various puzzle domains, such as 8-puzzles or Tower of Hanoi,
that are presented in the form of noisy, continuous visual depiction of the environment.

\latentplanner takes two inputs.
The first input is the \emph{transition input} $Tr$, a set of pairs of raw data.
Each pair $tr_i=(\before_i, \after_i) \in Tr$ represents a transition of the environment before and after some action is executed.
The second input is the \emph{planning input} $(i, g)$, a pair of raw data, which corresponds to the initial and the goal state of the environment.
The output of \latentplanner is a data sequence representing the plan execution that reaches $g$ from $i$.
While the original paper uses an image-based implementation (``raw data'' = images),
the type of data is arbitrary as long as it is compatible with NNs.

\begin{figure}[tb]
 \includegraphics[width=\linewidth]{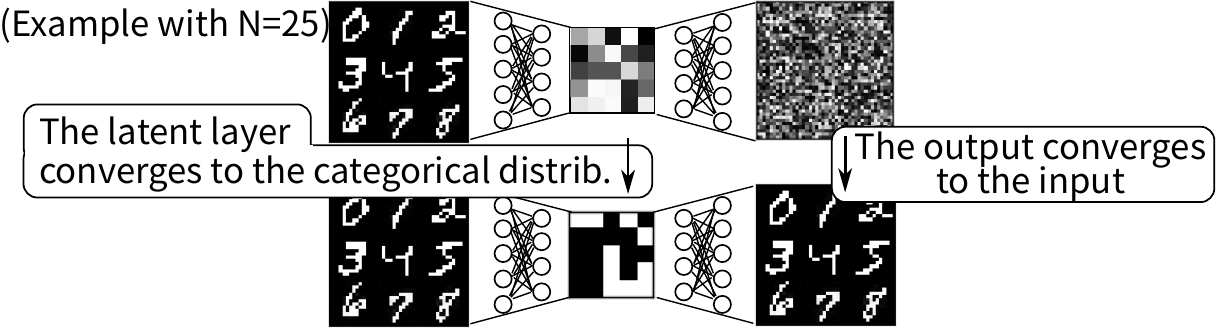}
 \caption{Step 1:
Train the State AutoEncoder by
 minimizing the sum of the reconstruction loss and the variational loss of Gumbel-Softmax.
As the training continues, the output of the network converges to the input images.
Also, as the Gumbel-Softmax temperature $\tau$ decreases during training,
the latent values approach either 0 or 1.}
 \label{sae}
\end{figure}

\latentplanner works in 3 phases.
In Phase 1, a \emph{State AutoEncoder} (SAE) (\refig{sae}) learns a bidirectional mapping between raw data (subsymbolic representation e.g., images)
 and propositional states (symbolic representation) from a set of unlabeled, random snapshots of the environment.
The trained SAE provides two functions:
\begin{itemize} 
\setlength{\itemsep}{-0.3em}
\item $b=Encode(r)$ maps an image  $r$ to a boolean vector $b$.
\item $\tilde{r}=Decode(b)$ maps a boolean vector $b$ to an image $\tilde{r}$.
\end{itemize}
After training the SAE from $\braces{\before_i, \after_i\ldots}$,
it applies $Encode$ to each $tr_i \in Tr$ and obtains $(Encode(\before_i),$ $Encode(\after_i))=$ $(s_i,t_i)=$ $\overline{tr}_i\in \overline{Tr}$,
the symbolic representations (latent space vectors) of the transitions.

In Phase 2, an Action Model Acquisition (AMA) method learns an action model (e.g., PDDL, successor function) from $\overline{Tr}$ in an unsupervised manner.
The original paper proposed two approaches: AMA$_1$ is an oracle which directly generates a PDDL without learning,
by allowing it to use the whole set of valid transitions as an oracle.
In contrast, AMA$_2$ approximates AMA$_1$ by unsupervised learning from examples.

In Phase 3, a planning problem instance is generated from the planning input $(i,g)$.
These are converted to the symbolic states by the SAE, and the symbolic planner solves the problem
combining $(i,g)$ and the generated action model.
For example, an 8-puzzle problem instance consists of an image of the start (scrambled) configuration of the puzzle ($i$), and an image of the solved state ($g$).

Since the intermediate states comprising the plan are SAE-generated latent bit vectors, the ``meaning'' of each state (and thus the plan) is not clear to a human observer.
However, in the final step, \latentplanner obtains a step-by-step visualization of the plan execution
by $Decode$'ing the latent bit vectors for each intermediate state.
The plans or the state transitions are validated based on the visualized result using a custom domain-specific validator implemented in Latplan.
This is because the intermediate latent representation is learned unsupervised
and is not directly verifiable through human knowledge.

\subsubsection{SAE as a Gumbel-Softmax Variational AutoEncoder}

The key concept of the SAE in \latentplanner is the use of Gumbel-Softmax \cite{jang2016categorical}
in the latent activation of the variational autoencoder~(VAE).
This allows the SAE to obtain a discretized binary representation, and \latentplanner uses this
discrete vector as the state representation for classical planning.
We briefly review the related literature below.

An autoencoder~(AE)~\cite{hinton2006reducing} is a feed-forward NN that consists of a pair of encoder and decoder networks, both of which are modeled by continuous functions.
For example, in Figure~\ref{sae}, the mapping from the leftmost image to the vector in the middle corresponds to the encoder, and the mapping from the middle to the rightmost image corresponds to the decoder.
AEs are trained so that the encoder maps a data point~(e.g., an image) into a low-dimensional latent space, and the decoder pulls the latent representation of the input back to the original data point.
Technically, they are trained by a \emph{backpropagation} algorithm so as to minimize the reconstruction loss, the distance between the input and the output measured by Euclidean distance or binary cross entropy.
Since the encoder is modeled by a continuous function, its latent representation is also continuous, which makes it challenging to integrate AEs with propositional reasoners.

A variational autoencoder (VAE) \cite{kingma2013auto} is a probabilistic variant of AEs, whose encoder and decoder are modeled by probabilistic distributions rather than deterministic functions.
For instance,
it obtains a discrete latent representation of the data by modeling the Bernoulli(=binary) random distribution with the encoder.
VAEs are trained by backpropagation with the help of the reparameterization trick, which makes random variables differentiable.
The reparametrization trick for the Bernoulli distribution is Gumbel-Softmax~\cite{jang2016categorical}.

A single Gumbel-Softmax unit in the Neural Network is able to model a categorical distribution with $M$ categories
which includes a Bernoulli distribution ($M=2$) as a special case.
Latplan uses $M=2$ for simplicity, and it does not affect the expressivity of the representation,
similar to the relation between SAS+ and the propositional representation.
\begin{align}
z_i &= \textbf{if}\ (i \ \text{is}\ \arg \max_i (g_i+\log \pi_i)) \ \textbf{then}\ 1\ \textbf{else}\ 0. \label{eq:gumbelmax} \\
z_i &= \text{Softmax}((g_i+\log \pi_i)/\tau).                   \label{eq:gumbelsoftmax}
\end{align}
The output of a single Gumbel-Softmax unit $GS(\boldsymbol{\pi}) = \mathbf{z}=(z_i)$ $(0\leq i \le M)$ is a one-hot vector representing $M$ categories, e.g.,
when $M=2$ the categories can be seen as $\braces{\text{false},\text{true}}$ and $\mathbf{z}=\parens{0,1}$ represents ``true''.
(Note: There is no explicit meaning assigned to each category.)
The input $\boldsymbol{\pi}=(\pi_i)$ is a class probability vector, e.g. $\parens{.2,.8}$.
Gumbel-Softmax is derived from Gumbel-Max technique \cite[\refeq{eq:gumbelmax}]{maddison2014sampling}
for drawing a categorical sample from $\boldsymbol{\pi}$
where $g_i$ is a sample drawn from
 $\text{Gumbel}(0,1) =-\log (-\log u)$ where $u=\text{Uniform}(0,1)$ \cite{gumbel1954statistical}.
Gumbel-Softmax approximates the argmax with a softmax to make it differentiable (\refeq{eq:gumbelsoftmax}).
``Temperature'' $\tau$ controls the magnitude of approximation, which is annealed to 0 by a certain schedule.
The output $\mathbf{z}$ converges to a discrete one-hot vector when $\tau\rightarrow 0$.

SAEs have $N$ Gumbel-Softmax units
to model an $N$-dimensional Bernoulli/boolean/propositional variable $\mathbf{b}=(b_n)$.
$N$ units produce a matrix $z_{nk}$ where $1\leq n \leq N$ and $k\in\braces{0,1}$ and
the boolean variables are retrieved by $b_n=z_{n1}$.

\section{Symbol Stability Problem}
\label{issues}

\begin{figure}[tb]
 \centering
 \includegraphics{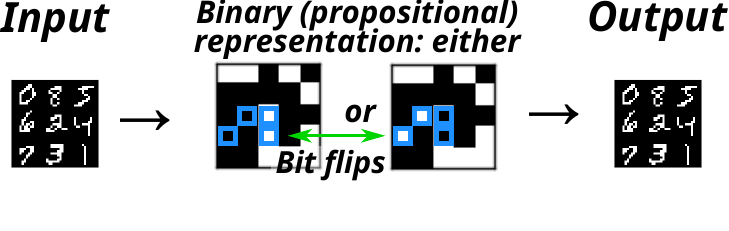}
 \includegraphics{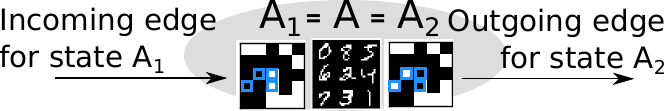}
 \caption{
(\textbf{Top}) Propositions found by SAEs may contain uninformative random bits
 that do not affect the output.
(\textbf{Bottom})
 The random variations of the propositional encoding could disconnect the search space.
}
 \label{unstable}
 \label{disconnected}
\end{figure}

The vanilla SAE in \latentplanner can map a visual observation of the environment to/from a set of propositional values.
An issue with the vanilla SAEs is that the class probability for the class ``true'' and the class ``false''
that is mapped to by the Gumbel-Softmax could be neutral at some neuron,
causing the value of the neuron to change frequently (\refig{unstable}).
The source of stochasticity is twofold.
The first source is the probabilistic distribution modeled by the encoder,
which introduces stochasticity and causes the propositions to change values even for the exact same inputs.
The second source is the stochastic observation of the environment which corrupts the input image.
When the class probabilities are almost neutral,
such a tiny variation in the input image may cause the activation to go across the decision boundary for each neuron,
causing the bit flips.
In contrast, humans still regard the corrupted image as the ``same'' image.

This stochastic behavior of the propositional representation
introduces several issues to the recipient symbolic systems such as classical planners.
Firstly,
search algorithms that run on the state space generated by the propositional vectors
are confused by many variations of the essentially identical real-world states.
It could visit the ``same'' real world state several times because
it could be encoded into different propositional vectors
which are not detected by the duplicate detection in the search algorithms (e.g. \astar).
This slows down the search by increasing the number of nodes that are reachable from the initial state.

Secondly, the state space could be disconnected due to such random variations (\refig{disconnected}).
Some states may be reached only via a single variation of the real world state and is not connected to
another propositional variation of the same real-world state.
In fact, in the appendix section in the Arxiv version of the original paper \cite{Asai2018},
the planning part used a so-called \emph{state augmentation} technique
which circumvents this issue by sampling states from the same image multiple times.

Thirdly, in order to reduce the stochasticity of the propositions, we encounter a hyperparameter tuning problem
which is costly when we train a large NN.
The neurons that behave randomly for the same or perturbed input do not affect the output,
i.e., they are unused and uninformative.
Unused neurons appear because the network has an excessive capacity to 
model the entire state space, i.e., they are surplus neurons.
Therefore, a straightforward way to reduce those neurons is to reduce the size of the latent space $N$.
On the other hand, if $N$ is too small, it lacks the capacity to represent the state space
and the SAE no longer learns to reconstruct the real world image.
As a result, we face a hyperparameter tuning problem: Surplus $N$ causes an unstable representation,
and insufficient $N$ makes the network hard to train.

Fundamentally,
the first two harmful effects are caused by breaking a critical feature of symbols, \emph{designation} \cite{newell1976computer},
that each symbol uniquely refers to an entity (referent, concept, meaning),
e.g., the referents of the symbols grounded by SAEs are the truth assignments.
If meaning of a symbol changes frequently and unexpectedly, the entire symbolic manipulation is fruitless
because the underlying symbols are not tied to any particular concept, and does not represent the real world.

Thus, for a symbol grounding procedure to produce a set of symbols for symbolic reasoning,
it is insufficient to find a set of symbols that are \emph{just able to} represent the environment;
It should find a \emph{stable} symbolic representation that \emph{uniquely} represents the environment.

\begin{defi}
\emph{Symbolic representation} of an environment is a set of symbols with referents
from which the environment can be reconstructed with sufficient accuracy.
\end{defi}

\begin{defi}
Symbolic representation is \emph{stable} when its referents are identical
for the same environment, under some equivalence relation (e.g., invariance, noise threshold).
\end{defi}

While NNs tend to achieve a robust performance on noisy data,
\textbf{the \emph{robust performance} and the stability of the representation are orthogonal.}
This is because the former exclusively deals with the output accuracy,
while the latter evaluates the quality of the \emph{latent} activations while maintaining the same output accuracy.
In fact, vanilla SAEs already achieve the almost perfect reconstruction accuracy
where the input and the output are indiscernible to human eyes
(e.g., in \refig{unstable}, the pixel value output is different from the input, but we cannot recognize it),
while they still exhibit instability.

The stability of the representation obtained by a NN depends
on
its inherent stochasticity during the runtime (as opposed to the training time) as well as
the stochasticity of the environment.
These observations indicate that \emph{any} symbol grounding system potentially suffers from 
the symbol stability problem.
As for the stochasticity of the environment,
in many real-world tasks, it is common to obtain stochastic observations
due to the external interference, e.g., vibrations of the camera caused by the wind.
As for the stochasticity of the network,
both
VAEs \cite{kingma2013auto,jang2016categorical,higgins2016beta} used in Latplan
and
GANs (Generative Adversarial Networks) \cite{goodfellow2014generative} used in Causal InfoGAN \cite{kurutach2018learning}
rely on sampling processes.

\section{Analyzing the State Autoencoder}
\label{analysis}

To obtain a deeper understanding of the mechanism that generates
unstable symbols in vanilla SAEs, we
analyzed its mathematical model and the public source code of Latplan (commit d48ee62).
We found that the source code differs from the original mathematical formulation of Gumbel-Softmax VAE (GS-VAE)
in that they are using an alternative loss function,
and we found that \emph{this very change} turned out to be essential to the success of their experiments
by suppressing the instability of the propositions.
In the following, we illustrate our finding using the idealized case where the encoder is the Bernoulli distribution,
although in reality, it is approximated by a Gumbel-softmax distribution.

Given a dataset $\mathcal{X}=\{\mathbf{x}^{(1)},\dots,\mathbf{x}^{(I)}\}$,
let $q(\mathbf{b} \mid \mathbf{x})$ and $p(\mathbf{x} \mid \mathbf{b})$ be the probabilities that the encoder and the decoder of GS-VAE respectively outputs the value $\mathbf{b}$ given $\mathbf{x}$ and $\mathbf{x}$ given $\mathbf{b}$,
where $\mathbf{x}\in \mathcal{X}$ corresponds to a visual observation of the environment, and $\mathbf{b}\in\{0,1\}^N$ corresponds to its latent representation.
In principle, GS-VAE is trained by minimizing the following objective function with respect to $p(\mathbf{x} \mid \mathbf{b})$ and $q(\mathbf{b}\mid \mathbf{x})$:
{
\relsize{-1}
\begin{align}
\label{vae-obj} \sum_{\mathbf{x}\in \mathcal{X}} \left(\mathbb{E}_{q(\mathbf{b}\mid \mathbf{x})}\left[-\log p(\mathbf{x}\mid \mathbf{b})\right] + \mathrm{KL}(q(\mathbf{b}\mid \mathbf{x}) \parallel p(\mathbf{b}))\right)
\end{align}
}
\\\noindent where $p(\mathbf{b}) = \prod_{n=1}^N p(b_n) = \prod_{n=1}^N \mathrm{Bern}(0.5)$ is the target, $N$-dimensional Bernoulli distribution with uniform probabilities
and $\mathrm{KL}(q || p)$ represents the Kullback-Leibler divergence from $p$ to $q$.
The first term in Eq.~\eqref{vae-obj} is the reconstruction loss, which measures the quality of the reconstructed data points,
and the second term regularizes the encoder by making the encoder $q(\mathbf{b}\mid \mathbf{x})$ closer to $p(\mathbf{b})$.
The second term is computed as
{
\relsize{-1}
\begin{align*}
 &{\mathrm{KL}}(q(\mathbf{b}\mid \mathbf{x}) \parallel p(\mathbf{b})) \\
= & - \sum_{n=1}^N \sum_{k\in\{0,1\}}q(b_{n}=k \mid \mathbf{x}) \log\frac{p(b_{n}=k)}{q(b_{n}=k \mid \mathbf{x})}\\
=& - \sum_{n=1}^N \sum_{k\in\{0,1\}}q(b_{n}=k \mid \mathbf{x}) \left(\log \frac{1}{2} - \log{q(b_{n}=k \mid \mathbf{x})}\right)\\
=& - \sum_{n=1}^N H_q(b_n\mid \mathbf{x}) + \mathrm{const.} = -H_q(\mathbf{b} \mid \mathbf{x}) + \mathrm{const.},
\end{align*}
}
where $H_q(\mathbf{b}|\mathbf{x})$ is the entropy of $\mathbf{b}$ given $\mathbf{x}$ under $q$.

\subsubsection{Entropy Regularization in Latplan}

We found that the loss computation in the Latplan code has the \emph{opposite sign} on the KL divergence,
i.e., it is \emph{maximizing} the KL divergence instead of minimizing it.
The system works
because maximizing the KL divergence corresponds to minimizing the entropy of $q$, thus finding a stable representation.
This is natural considering the nature of the original GS-VAE:
The original loss function of the GS-VAE tries to make the latent distribution closer to the fair random Bernoulli distribution $p(\mathbf{b})$
that takes 0 or 1 with the equal probability, i.e., \emph{as random as possible},
which is, in fact, opposite from the concept of stability.
Instead, Latplan has a negated loss, which resulted in
maximizing the KL divergence and making the representation \emph{less random}.

The resulting loss function implemented in Latplan is therefore as follows:
{
\relsize{-1}
\begin{align} 
\nonumber  &\brackets{\text{rec loss}}-{\mathrm{KL}}(q||p)                                    \\
\nonumber =&\brackets{\text{rec loss}}+{\mathrm{KL}}(q||p) - 2 {\mathrm{KL}}(q||p)          \\
\label{loss+ent} =&\brackets{\text{\small the original GS VAE loss}}      + 2 H_q(\mathbf{b} \mid \mathbf{x}).
\end{align}
}
Since the entropy measures the randomness of the random variables,
the extra entropy term in Eq.~\eqref{loss+ent} regularizes the network by penalizing the unstable representation.
In the later sections, we empirically show that the original loss function (Eq.~\eqref{vae-obj}) for GS-VAE results in
a much higher instability compared to the GS-VAE with the Entropy regularization (Eq.~\eqref{loss+ent}).

\subsubsection{Removing the Run-Time Stochasticity}
\label{argmax}

Another improvement we made from the original approach in Latplan is that we can
disable the stochasticity of the network while performing the planning.
After the training is finished, we replace the Gumbel-softmax activation with
a pure argmax of class probabilities, which makes the network fully deterministic:
\[
 z_{i} = \textbf{if}\ (i\ \text{is} \arg \max_i (\log \pi_{i}))\ \textbf{then}\ 1\ \text{else}\ 0.
\]
This technique reduces the inherent stochasticity of the network.

\section{Zero-Suppressed State AutoEncoder}
\label{zsae}

In Latplan, Entropy Regularization was the key to address the symbol stability (see \refsec{evaluation})
while it was only accidentally introduced.
In this context, a natural next step toward obtaining the more stable symbols is to introduce a new regularization
for the propositional representation.
We propose Zero-Suppressed State AutoEncoder (ZSAE),
an SAE with an additional regularization designed for the discrete representation that we call \emph{zero-suppression}.
Its fundamental idea is to penalize the
true propositions in the latent layer so that no propositions unnecessarily flip to true at random
while preserving the propositions that are absolutely necessary for maintaining the reconstruction accuracy.
The resulting loss is 
\emph{asymmetric} to a particular label $k=1$ (true):
\begin{align*}
 \brackets{\text{loss}} = & \brackets{\text{\small vanilla SAE loss}} + \alpha \sum_n \sum_{k\not=0} z_{nk} 
\end{align*}
where $\alpha$ is the magnitude of regularization.
This formulation takes a general form
that also covers the multi-valued SAS+ representation with $k\geq 2$.
In principle, this method could also be used for a SAS+ representation,
but we focus on the binary representation in this paper.

One additional advantage of the ZSAE is that
several neurons are completely deactivated, i.e. they always take the value of zero
and can be pruned afterward to reduce the network size,
similar to Zero-Suppressed Decision Diagrams \cite{minato1993zero}.
Unlike traditional NN compression methods \cite{cheng2017survey}, it does not suffer from
accuracy degradation because the activations are discrete and therefore no additional retraining is required.
In the continuous cases, even the minuscule activations could be amplified by the weights and significantly affect the
later pipelines of the neural networks.
Our method for the discrete representations complements
the prior work for the continuous ones.

Assume the propositional layer $z_{nk}$ is connected to the next layer of $L$ neurons
by a fully-connected network $h_l=\sigma\parens{\sum_{n=1}^N \sum_{k\in\braces{0,1}} W_{nkl}z_{nk}+B_l}$
with weights $W_{nkl}$, biases $B_l$, and a nonlinear activation $\sigma$.
($1\leq l \leq L$.)
When we assume that $z_{n0}=1$ and $z_{n1}=0$ always holds ($b_n=0$ for all inputs),
we can prune the index $n$
by adding $W_{n0l}$ to $B_l$ and removing $W_{nkl}$ for $\forall k\in\braces{0,1}$, $\forall l \in \braces{1,...L}$,
which removes $2L$ float values for each zero-suppressed bit $b_n$.
Therefore, if we remove $\Delta N$ bits from the latent space, it removes $2L\Delta N$ float values from the network.
We can similarly prune the weights $W'$ from the previous layer of $M$ neurons.

\textbf{Implementation Note.} Regardless of $\alpha$,
the regularization tends to be too strong near the beginning of the training.
In practice, we set $\alpha=0$ until 1/3 of the total epochs.
We confirmed that gradually increasing $\alpha$ also works,
but we did not use it in the later experiments.

\section{Empirical Evaluation}
\label{evaluation}

We evaluated various SAE implementations across 5 different
image domains depicting 8-puzzles or Lights Out puzzle game \cite{lightsout}.
Each training takes at most 30 minutes.

\textbf{MNIST 8-puzzle}
is an image-based version of the 8-puzzle, where tiles contain hand-written digits (0-9) from the  MNIST database \cite{lecun1998gradient}.
Valid moves in this domain swap the ``0'' tile  with a neighboring tile, i.e., the ``0'' serves as the ``blank'' tile in the classic 8-puzzle. 
The \textbf{Scrambled Photograph 8-puzzle (Mandrill, Spider)} cuts and scrambles real photographs, similar to the puzzles sold in stores).
\textbf{LightsOut} \cite{lightsout} is
a game where a grid of lights is in some on/off configuration ($+$: On),
and pressing a light toggles its state as well as the states of its neighbors.
The goal is all lights Off.
\textbf{Twisted LightsOut} distorts the original LightsOut game image by a swirl effect.

\subsection{The Quality of the Latent Representation}

We compare the quality of the latent representation
produced by the ZSAE and the vanilla SAE.

\subsubsection{State Variance}

The first metric we evaluated is the bit-wise variance of the state encoding for the same/similar input,
which directly measures the stability of the representation.
We trained several SAEs for each domain with the different latent layer sizes (numbers of propositions) $N$
and then evaluated the variance.
In all experiments below,
we randomly generated 100 images with a domain-specific generator for each puzzle domain,
then encoded each of them with the SAE 100 times.
We measured the variance of the propositions, i.e. the variance of latent activations (0 or 1)
across 100 encoding trials of the same image.
We then took the mean of the variances over the entire propositions.

We evaluated three versions of the SAE:
(1) NG-SAE, an SAE trained with the original GS-VAE loss function as discussed in \refsec{analysis}, and
(2) Vanilla SAE in the original paper of Latplan \cite{Asai2018} and the Github source code,
(3) Zero-Suppressed SAE (ZSAE).

The first thing we tested is to replace the Gumbel-softmax activation with a deterministic argmax function
after the training (\refsec{argmax}).
In all SAEs, this reduced the variance to 0 for a single input because all networks become deterministic.
We omit the results due to space because it is rather obvious.
In the following experiments, we always replace the activation function with argmax during testing.

We next measured the variance of SAEs in a noisy setting, where
we perturbed the input image by Gaussian noise for each of the 100 trials of the same image.
\reftbl{tab:stability} (first columns) indicates that
the propositions made by NG-SAE are highly random,
while the entropy regularization in the vanilla SAE suppresses the stochastic behavior to some extent.
ZSAE further reduces the variance and achieves the most stable representation.
The effect of SAE$\rightarrow$ZSAE was typically 2-3 and up to 4 orders of magnitude ($2.5e-4\rightarrow 4.5e-8$),
much stronger than that of Entropy Regularization (NGSAE$\rightarrow$SAE, 1 to 2 orders of magnitude).
Due to the poor performance of NG-SAE, we do not study it any further in the later experiments.

\begin{table*}[tbp]
 \centering
 \resizebox{.95\textwidth}{!}{
 \begin{tabular}{|r|*{17}{c|}}
       & \multicolumn{6}{c|}{Mean variance over bits (with noisy images)}
       & \multicolumn{5}{c|}{Effective bits}
       & \multicolumn{6}{c|}{Mean Square Error (MSE)}
  \\
$N=$ & \multicolumn{3}{c|}{100} & \multicolumn{3}{c|}{1000}
     & \multicolumn{2}{c|}{100} & \multicolumn{2}{c|}{1000}
     & Optimal
     & \multicolumn{2}{c|}{100} & \multicolumn{2}{c|}{1000} & \multicolumn{2}{c|}{36}
  \\
domain    & NG-SAE & SAE    & ZSAE            & NG-SAE & SAE    & ZSAE            & SAE & ZSAE        & SAE  & ZSAE & Encoding & SAE     & ZSAE   & SAE    & ZSAE   & SAE    & ZSAE           \\ 
MNIST     & 8.4e-2 & 8.6e-3 & \textbf{3.7e-6} & 5.3e-2 & 2.2e-4 & \textbf{1.1e-7} & 100 & 51          & 1000 & 68   & 18.4     & $<$1e-4 &$<$1e-4 &$<$1e-4 &$<$1e-4 &$<$1e-4 &\uline{9.1e-3}  \\ 
Mandrill  & 1.1e-3 & 8.3e-4 & \textbf{3.0e-5} & 4.2e-4 & 2.5e-4 & \textbf{4.5e-8} & 100 & 46          & 1000 & 182  & 18.4     & 3.0e-4  &2.8e-4  &2.1e-4  &2.3e-4  &2.0e-4  &{3.2e-4}        \\ 
Spider    & 8.5e-4 & 4.9e-4 & \textbf{6.3e-6} & 2.3e-4 & 4.2e-4 & \textbf{7.3e-7} & 100 & 49          & 1000 & 200  & 18.4     & 2.7e-4  &2.2e-4  &3.1e-4  &2.8e-4  &$<$1e-4 &\uline{2.8e-2}  \\ 
L-Out     & 9.0e-3 & 2.0e-4 & \textbf{3.9e-6} & 8.1e-3 & 1.4e-4 & \textbf{7.5e-6} & 100 & \textbf{16} & 1000 & 66   & 16       & $<$1e-4 &$<$1e-4 &$<$1e-4 &$<$1e-4 &2.9e-4  &{2.8e-4}        \\ 
Twisted   & 1.0e-2 & 7.1e-4 & \textbf{5.1e-6} & 1.0e-2 & 4.5e-4 & \textbf{1.6e-7} & 100 & \textbf{16} & 1000 & 49   & 16       & $<$1e-4 &$<$1e-4 &$<$1e-4 &$<$1e-4 &$<$1e-4 &\uline{5.7e-3}  \\ 
\end{tabular}
}
 \caption{\textbf{Representation characteristics.}
Results comparing the NG-SAE, vanilla SAE and ZSAE ($\alpha$=0.7).
(\tb{Left}) Comparing the representation variance over 100 randomly generated images encoded 100 times with Gaussian noise added each time.
(\tb{Middle}) The number of bits that ever turns true when encoding the entire state space.
 In LightsOut and Twisted, ZSAE($N$=100) finds an optimal, 16-bit representation of the 4x4 puzzle.
(\tb{Right}) Mean Square Error for the test data.
 }
\label{tab:stability}
\end{table*}

\subsubsection{Effective Size of the Representation}

Next, \reftbl{tab:stability} (middle columns) shows that the number of effective bits,
i.e. the number of propositions that \emph{ever} change their values over all states, is low in ZSAE, showing that
ZSAE obtained a more compressed, compact representation of the input.
In MNIST, the numbers are comparable between ZSAEs with $N$=100,1000,
which shows that the network is able to find an encoding of almost the same size
regardless of the size of the latent layer (upper bound of the size of
propositions), reducing the need for hyperparameter tuning.
In LightsOut and Twisted, ZSAE even finds the 16bit optimal representation for the 4x4 light grids.

\subsection{Output Accuracy}

Regularization in general works by restricting the neural network
to achieve some desirable property, e.g., most commonly for suppressing the overfitting \cite[chap.5]{Goodfellow-et-al-2016}, but in our case
for improving the stability. Thus, as a result of the restricted expressive capacity, the output accuracy may be degraded
when the regularization is too strong.

Thus we next tested if the zero-suppression affects the output accuracy.
In \reftbl{tab:stability} (right columns)
we show the Mean Square Error between the input and the output
for 100 randomly generated images.
The results indicate that the zero-suppression does not significantly affect the output accuracy for $N$=100,1000.
However, for $N$=36 (a parameter tuned for vanilla SAEs to have the least variance), the zero-suppression
harmed the accuracy because the network is already small and the further penalty affected the training.
In other words, it is better to combine ZSAE with an overcapacity network,
since ZSAE then automatically compresses the representation.

\subsubsection{Hyperparameter Sensitivity}

\begin{table}[htb]
 \centering
 \resizebox{.95\columnwidth}{!}{
 \begin{tabular}{|r|*{8}{c|}}
     & \multicolumn{8}{c|}{Mean variance over bits (with noisy images)} \\
     & \multicolumn{3}{c|}{SAE} 
     & \multicolumn{2}{c|}{ZSAE($\alpha$=0.7)} 
     & \multicolumn{3}{c|}{ZSAE($N$=100)}
  \\
$N=$      &36     & 100    & 1000   & 100    & 1000   & $\alpha=$0.2 & 0.5    & 0.7    \\
MNIST     &1.8e-4 & 8.6e-3 & 2.2e-4 & 3.7e-6 & 1.1e-7 & 4.5e-7       & 0.0e+0 & 3.7e-6 \\
Mandrill  &2.9e-4 & 8.3e-4 & 2.5e-4 & 3.0e-5 & 4.5e-8 & 2.3e-6       & 3.4e-6 & 3.0e-5 \\
Spider    &5.8e-6 & 4.9e-4 & 4.2e-4 & 6.3e-6 & 7.3e-7 & 4.6e-6       & 8.3e-6 & 6.3e-6 \\
L-Out     &2.5e-6 & 2.0e-4 & 1.4e-4 & 3.9e-6 & 7.5e-6 & 1.1e-4       & 1.5e-5 & 3.9e-6 \\
Twisted   &2.1e-5 & 7.1e-4 & 4.5e-4 & 5.1e-6 & 1.6e-7 & 8.8e-5       & 4.2e-6 & 5.1e-6 \\
\end{tabular}
}
 \caption{Sensitivity of ZSAE to the hyperparameter $\alpha,N$ compared to that of SAE to the hyperparameter $N$.}
 \label{sensitivity}
\end{table}

We tested the sensitivity of ZSAE to its new hyperparameter $\alpha$ (\reftbl{sensitivity}).
The purpose of this experiment is to show that ZSAEs are less sensitive to the choice of both $N$ and $\alpha$,
while vanilla SAEs are sensitive to $N$.
We compared the state variances under noise
between the vanilla SAE and the ZSAE with $\alpha=$0.2,0.5,0.7, $N$=100,1000.
We also added $N$=36 as the best case for the vanilla SAE.
We observed that the ZSAEs achieve the comparable state variance with various $\alpha$ and $N$,
while the vanilla SAEs are significantly affected by $N$.
For instance,
the variance for the vanilla SAE with $N$=100 is up to two orders of magnitude larger than that for the SAE with $N$=36.
In contrast,
the worst case for the ZSAEs is 1.1e-4 on Twisted with $(N,\alpha)$=(100,0.2),
which is still better than the vanilla SAE with the same $N$.
They also have similar reconstruction loss (MSE) and the effective bits.
Therefore, we conclude that while ZSAE introduces an additional parameter,
the selection of $N$ and $\alpha$ is easy compared to the selection of $N$ in SAE.

\subsection{Planner Performance}

Next, we compared the success ratio of \latentplanner with various parameters.
We tested both AMA$_1$ and AMA$_2$ proposed in \cite{Asai2018} as the Action Model Acquisition (AMA) methods.
Each domain has 60 problem instances each generated by a random walk from
the goal state. 60 instances consist of 30 instances generated by 7-steps random walks
and another 30 by 14 steps. 30 instances consist of 10 instances whose images are corrupted by Gaussian noise,
10 with salt/pepper noise and another 10 without noise.
It is important to note that \emph{the noiseless instances do not have the external stochasticity},
one of two sources of instability.

We first tested AMA$_1$, an oracular, idealistic AMA that does not incorporate machine learning,
and instead generates the entire propositional state transitions from the entire image transitions.
The purpose we test an impractical AMA$_1$ method is
to separate the effect of a better state representation achieved by ZSAE
and that of the learning procedure in AMA$_2$ that learns the state transitions and the action rules.
As a classical planner, we used FastDownward \cite{Helmert04} with \astar, blind heuristics in order to
remove the effects of the heuristic functions. We ran the solver with no runtime/memory restrictions.

The results in \reftbl{tab:ama1} (left) show that
ZSAEs have \textbf{tripled} the score of SAE ($14\rightarrow43, 6\rightarrow48, 6\rightarrow33$)
under external stochasticity (Gaussian noise).
Regarding the hyperparameter sensitivity,
ZSAE showed a robust performance across $N$=36,64,100
while SAE achieves a good performance only in a single parameter $N$=36.
Due to the small size of the state space compared to the usual classical planning benchmark domains,
the search finishes in a fraction of a second, which means \textbf{the failure is due to graph disconnectedness at the initial/goal nodes}.
ZSAE tends to fail in $N$=36 in LightsOut and Twisted because of the higher reconstruction error
discussed in the previous subsection.

\begin{table*}[tb]
\centering
\resizebox{.95\textwidth}{!}{
\begin{tabular}{|l|*{6}{*{3}{r}|}|*{6}{*{3}{r}|}}
 \hline
 & \multicolumn{18}{c||}{\textbf{AMA$_1$ results}} & \multicolumn{18}{c|}{\textbf{AMA$_2$ results}}
 \\
 & \multicolumn{6}{c|}{Gaussian ($\sigma$=0.6)}
 & \multicolumn{6}{c|}{Salt/Pepper ($p$=0.12)}
 & \multicolumn{6}{c||}{No noise}
 & \multicolumn{6}{c|}{Gaussian ($\sigma$=0.6)}
 & \multicolumn{6}{c|}{Salt/Pepper ($p$=0.12)}
 & \multicolumn{6}{c|}{No noise}
 \\
 & \multicolumn{3}{c|}{SAE} & \multicolumn{3}{c|}{ZSAE}
 & \multicolumn{3}{c|}{SAE} & \multicolumn{3}{c|}{ZSAE}
 & \multicolumn{3}{c|}{SAE} & \multicolumn{3}{c||}{ZSAE}
 & \multicolumn{3}{c|}{SAE} & \multicolumn{3}{c|}{ZSAE}
 & \multicolumn{3}{c|}{SAE} & \multicolumn{3}{c|}{ZSAE}
 & \multicolumn{3}{c|}{SAE} & \multicolumn{3}{c|}{ZSAE}
 \\
$N=$
 & {36} & {64} & {100}  & {36} & {64} & {100} 
 & {36} & {64} & {100}  & {36} & {64} & {100} 
 & {36} & {64} & {100}  & {36} & {64} & {100} 
 & {36} & {64} & {100}  & {36} & {64} & {100} 
 & {36} & {64} & {100}  & {36} & {64} & {100} 
 & {36} & {64} & {100}  & {36} & {64} & {100} 
 \\
\hline
MNIST          & 0       & 0  & 0       & \tb{17} & 0       & 0       & 17      & 5  & 0       & \tb{20} & \tb{10} & \tb{17} & 20  & 20  & 20      & 20      & 20      & 20  
               & 8       & 8  & 4       & \tb{20} & \tb{14} & \tb{10} & 18      & 6  & 0       & \tb{20} & \tb{9}  & \tb{17} & 18  & 13  & 5       & \tb{20} & \tb{19} & \tb{18} \\
Mandrill       & \tb{4}  & 3  & 0       & 0       & \tb{6}  & \tb{5}  & 16      & 17 & 8       & \tb{20} & \tb{20} & \tb{20} & 20  & 20  & 20      & 20      & 20      & 20  
               & \tb{13} & 10 & 3       & 4       & \tb{13} & \tb{12} & 18      & 13 & 6       & 18      & \tb{19} & \tb{19} & 18  & 13  & 10      & 18      & \tb{19} & \tb{20} \\
Spider         & 8       & 3  & \tb{6}  & \tb{12} & 3       & 5       & 20      & 7  & 16      & 20      & \tb{20} & \tb{20} & 20  & 20  & 20      & 20      & 20      & 20  
               & 13      & 11 & \tb{17} & \tb{17} & \tb{18} & 16      & 17      & 10 & 15      & \tb{18} & \tb{20} & \tb{18} & 17  & 14  & 15      & 17      & \tb{20} & \tb{18} \\
L-Out          & 1       & 0  & 0       & \tb{10} & \tb{19} & \tb{20} & \tb{20} & 13 & 18      & 12      & \tb{19} & \tb{20} & 20  & 20  & 20      & 20      & 20      & 20  
               & 10      & 0  & 0       & \tb{20} & \tb{20} & \tb{12} & 20      & 20 & \tb{20} & 20      & 20      & 18      & 20  & 19  & \tb{20} & 20      & \tb{20} & 18 \\
Twisted        & 1       & 0  & 0       & \tb{4}  & \tb{20} & \tb{3}  & \tb{17} & 12 & 6       & 12      & \tb{20} & \tb{17} & 20  & 20  & 20      & 20      & 20      & 20  
               & 3       & 0  & 0       & \tb{10} & \tb{16} & \tb{12} & 19      & 17 & \tb{20} & \tb{20} & \tb{20} & 15      & 20  & 20  & \tb{20} & 20      & 20      & 15 \\ \hline
\textbf{Total} & 14      & 6  & 6       & \tb{43} & \tb{48} & \tb{33} & \tb{90} & 54 & 48      & 84      & \tb{89} & \tb{94} & 100 & 100 & 100     & 100     & 100     & 100 
               & 47      & 29 & 24      & \tb{71} & \tb{81} & \tb{62} & 92      & 66 & 61      & \tb{96} & \tb{88} & \tb{87} & 93  & 79  & 70      & \tb{95} & \tb{98} & \tb{89} \\
\hline
\multicolumn{4}{|l|}{Total (ZSAE, $\alpha$=0.2)} & \tb{21} & \tb{10} & 2       & & & & 75 & \tb{88} & \tb{68} & & & & 100 & 100     & 99  &
\multicolumn{3}{|l|}{}                           & 46      & 29      & \tb{33} & & & & 76 & \tb{72} & \tb{71} & & & & 76  & \tb{86} & \tb{81} \\
\multicolumn{4}{|l|}{Total (ZSAE, $\alpha$=0.5)} & \tb{15} & \tb{33} & \tb{12} & & & & 77 & \tb{94} & \tb{89} & & & & 100 & 100     & 100 &
\multicolumn{3}{|l|}{}                           & 36      & \tb{41} & \tb{27} & & & & 56 & \tb{81} & \tb{65} & & & & 59  & \tb{92} & \tb{72} \\
\hline
\end{tabular}
}
\caption{
\textbf{Planning Results.}
(\textbf{Left})
The number of instances successfully solved by Latplan using AMA$_1$ (oracular method)
for comparing the performance of Z/SAE.
Better results among the same configuration of ZSAE/SAE are highlighted in \tb{bold}.
SAEs degrade performance as the surplus capacity produces more unstable propositions
and is better than ZSAE only when tuned to $N$=36.
ZSAEs are robust on the different $N$ and tend to solve more problems than the vanilla SAEs.
(\textbf{Right})
The numbers of instances solved under 180 sec using AMA$_2$ unsupervised learning method for Action Model Acquisition.
Results indicate that ZSAEs are robust on the different hyperparameters and tend to achieve better performance than vanilla SAEs.
}
\label{tab:ama1}
\label{tab:ama2}
\end{table*}

Next, we compare the planning performance of Z/SAE with AMA$_2$,
a NN-based AMA model that learns from the example state transitions.
It consists of two networks:
(1) Action AutoEncoder (AAE), an autoencoder that learns to cluster the state transitions into a finite number of action labels.
It learns to reconstruct the input propositional successor state $t$ in the output $\hat{t}$.
Its latent layer represents an action label $a$, which is a single Gumbel-Softmax-activated unit of $A$ categories,
where $A$ is the upper bound of the number of action labels.
In addition, every layer of AAE is concatenated with the propositional vector of the current state $s$ as the \emph{secondary} input,
which turns the standard AE formulation of $f(t)=a,\ g(a)=\hat{t}$ into $f(t,s)=a,\ g(a,s)=\hat{t}$,
where $g$ can be interpreted as a function that \emph{applies} an action $a$ to $s$ and obtains $\hat{t}$.
(2) Action Discriminator (AD), a binary classifier modeling the action precondition, which takes $(s,t)$ and returns a boolean.
Combining AAE and AD yields a successor function that can be used for graph search algorithms.
Networks in AMA$_2$ are trained with the same hyperparameters used in the original paper \cite{Asai2018}.
\reftbl{tab:ama2} (right) shows that ZSAEs \textbf{doubled} ($29\rightarrow81, 24\rightarrow62$) the success ratio over vanilla SAEs
under noise.
The gap is not as large in the noiseless scenario, but note that
the scenario is unrealistic: Virtually all real-world data contain various forms of noise/perturbation.

\begin{figure}[tb]
 \centering
 \includegraphics[width=0.49\linewidth]{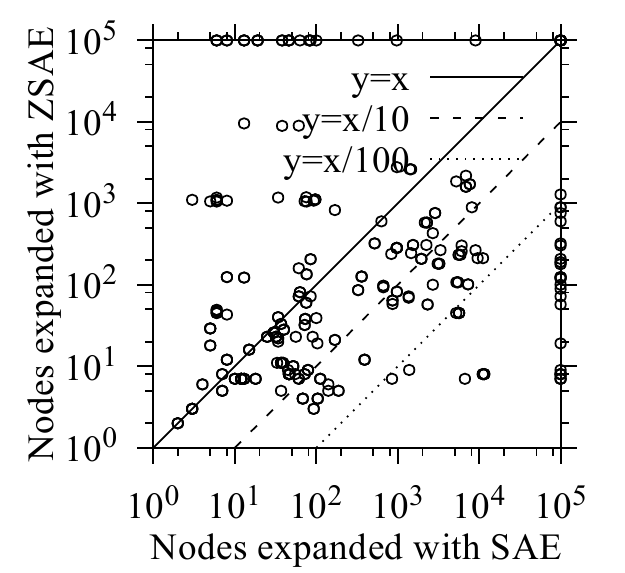}
 \includegraphics[width=0.49\linewidth]{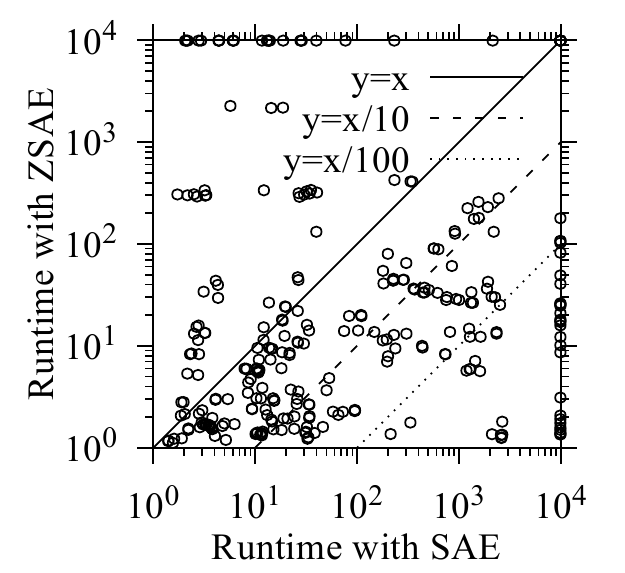}
 \caption{
Double-logarithmic plot of the number of states expanded (left) / time spent (right) by
\astar with goal-count heuristics,
for instances successfully solved under 1 hour by both SAE+AMA$_2$ and ZSAE($\alpha$=0.7)+AMA$_2$, both with $N$=100, all domains.
In both figures,
$x$-axes represent the SAE and $y$-axes represent the ZSAE.
Unsolved instances are shown on the borders.
We observe that the search effort tends to be larger with the SAE than with the ZSAE.
Opposite cases are generated due to
the tie-breaking difference and
the random disconnectedness caused by the SAE.
}
 \label{fig:ama2-statistics}
\end{figure}

Finally, we compared the search statistics between SAE+AMA$_2$ and ZSAE+AMA$_2$
in order to measure another harmful effect of the unstable propositions
that they \textbf{confuse the duplicate detection of search algorithms and increase the search effort}.
This effect cannot be measured with AMA$_1$
because it creates PDDL/SAS instances based on the fixed set of
input images and both SAE/ZSAE are deterministic (using argmax),
therefore the search graph contains a single node for a single image.
This is not the case with AMA$_2$ because the NNs generate the successor states on the fly.
We measured the number of node expansions (left) and the runtime (right)
on the problems successfully solved by both SAE+AMA$_2$ and ZSAE+AMA$_2$ (\refig{fig:ama2-statistics}).
In order to gather the sufficient number of data points for comparing the search statistics of SAE and ZSAE,
we extended the maximum runtime limit of 1 hour
and reduced the input noise so that SAE can solve a sufficient number of problem instances.
The parameters for the Gaussian noise and the salt/pepper noise applied to the input
are $\sigma=0.3$ and $p=0.06$, respectively, compared to $\sigma=0.6$ and $p=0.12$ in \reftbl{tab:ama2}.
In this setting, the gap between ZSAE and SAE narrowed: ZSAE solved 266 instances and SAE solved 262 instances in total.
The plots support our claim that
the randomness in the state encoding of vanilla SAE confuses the duplicate detection and
increase the search effort.
ZSAE resulted in a smaller node expansion in 172 out of 238 instances solved by both SAE and ZSAE,
and in a shorter runtime in 191 out of 238 instances.

\subsection{Pruning Inactive Nodes from ZSAE}

We discuss the amount of memory reduction possible by the pruning on the nodes
that have constant activations of 0. As we saw from \reftbl{tab:stability},
vanilla SAEs do not have such propositions (all bits are effective).

Representing a fully-connected
network between two layers of $L$ and $N$ nodes requires $(L+1)N$
weights ($+1$ for the bias).
In the network we used,
both the previous and the succeeding layer of the latent propositional layer have 1000 nodes.
In Gumbel-softmax, each proposition corresponds to 2 neurons.
Thus, in the case of ZSAE with $N$=1000 applied to MNIST puzzles,
the representation is compressed down to 68 effective bits and
the weights are reduced by $(1000+1)\times (2\cdot 1000 - 2\cdot 68)$=1865864 for the previous layer,
and $((2\cdot 1000+1)-(2\cdot 68+1))\times 1000 $=1864000 for the succeeding layer.
This number is huge compared to the convolutional weights (3x3, 16 channels, thus 144 float values each) in the upper layers.
The total number of weights in the network is reduced from 8376278 to 5578414 (44\% reduction).
We do not show the entire results because the results are straightforward:
The numbers can be similarly calculated from the numbers on \reftbl{tab:stability}.

\section{Related Work}

The proposed method has some similarities to the common technique called $\ell_1$ regularization \cite[Sparse AE]{Goodfellow-et-al-2016},
which is applied to the continuous activations of the neurons or the training weights
and achieves a sparse representation where the continuous values become closer to zero.
$\ell_1$ regularization is applied to all neurons in the same layer
while the proposed zero-suppression is applied to a subset of neurons representing a particular discrete value.
Moreover, regularization is traditionally applied in order to suppress overfitting and improve predictive performance
while our focus is rather on the stability of the representation.

A possible interpretation of this approach 
is that
this regularization is working as a model prior corresponding to \emph{closed-world assumption} \cite[CWA]{reiter1981closed},
which assumes that all propositions are false
when they are unknown to a Knowledge Base (KB) or cannot be proven from it.
With CWA, KBs no longer have to explicitly store the false propositions,
just as the ZSAE can prune the constant-0 nodes.

The search space generated by \latentplanner was shown to be compatible
to an existing Goal Recognition system \cite{amado2018goal,amado2018goalb}.
Another recent approach replacing SAE/AMA$_2$ with InfoGAN \cite{kurutach2018learning}
has no explicit mechanism for improving the stability of the binary representation.
Other methods for generating a symbolic model from the environment \cite{YangWJ07,CresswellMW13,MouraoZPS12}
require symbolic or near-symbolic, structured inputs.
Konidaris et al. (\citeyear{KonidarisKL18}) generates a symbolic state space from
the low-level sensor inputs but requires the high-level action symbols.
Our approach complements the usability of these approaches by providing more stable symbols.

Previous work in Learning from Observation \cite{BarbuNS10,Kaiser12},
which could produce propositions from the observations (unstructured input)
with the help of hand-coded symbol extractors (e.g. \emph{ellipse detectors} for tic-tac-toe),
typically assume 
a perfect indoor environment with little noisy interference,
therefore do not have to address the stability of the symbols.

\section{Conclusion}
\label{conclusion}

In this paper, we provided a formal analysis of the State AutoEncoder
(SAE) neural network (NN) in \latentplanner \cite{Asai2018} to
understand its success and 
the issue of its \emph{unstable propositions}.
In the analysis, we identified that the SAE accidentally
uses a different formulation of Gumbel-Softmax VAE \cite{jang2016categorical}
that we named \emph{Entropy Regularization} which stabilizes the truth values.

To further improve the stability,
we introduced Zero-Suppressed State AutoEncoder (ZSAE) which
improves the vanilla SAE by
minimizing the number of true propositions in the representation.
ZSAE improves the success rate and the efficiency of planning performed on
the generated state space and
also removes the need for aggressive hyperparameter tuning.
Moreover, 
ZSAE makes it possible to prune some neurons without accuracy degradation
when their activations are constantly zero, 
similar to Zero-Suppressed Decision Diagrams \cite{minato1993zero} and
the knowledge bases with closed world assumption \cite{reiter1981closed}.

As a meta-level contribution,
we generalized the problematic behavior of the unstable propositions
into a \emph{Symbol Stability Problem} (SSP), a subproblem of symbol grounding.
We identified two sources of stochasticity which can introduce the instability:
The inherent stochasticity of the network and
the external stochasticity from the observations.
This suggests that
SSP is an important problem that applies to any modern NN-based symbol grounding process
that operates on the noisy real-world inputs and
performs a sampling-based, stochastic process (e.g. VAEs, GANs) that are gaining popularity in the literature.
Thus, characterizing the aspect of SSP would help the process of designing a planning system operated on the real world input.
An interesting avenue for future work is to extend our approach to InfoGAN-based
discrete representation of the environment \cite{kurutach2018learning}.

\fontsize{9pt}{10pt}            
\selectfont

\bibliographystyle{aaai}
\end{document}